\documentclass{article}
\usepackage{cite}
\usepackage{spconf,amsmath,graphicx}
\usepackage{amssymb,booktabs,multirow,hyperref} 

\usepackage{fancyhdr}

\fancypagestyle{firststyle}
{
  \fancyhf{}
\fancyhead[R]{\thepage}
\fancyfoot[C]{\footnotesize © 2025 IEEE. Personal use of this material is permitted. Permission from IEEE must be obtained for all other uses, in any current or future media, including reprinting/republishing this material for advertising or promotional purposes, creating new collective works, for resale or redistribution to servers or lists, or reuse of any copyrighted component of this work in other works.
}

}

\name{
  \begin{tabular}{cccc}
    Alper Bahcekapili$^{1}$ & Duygu Arslan$^{1}$ & Umut Ozdemir$^{1}$ & Berkay Ozkirli$^{1}$ \\
    Emre Akbas$^{1,2}$ & Ahmet Acar$^{1}$ & Gozde B. Akar$^{1}$ & Bingdou He$^{3}$ \\
    Shuoyu Xu$^{3}$ & Umit Mert Caglar$^{1}$ & Alptekin Temizel$^{1}$\thanks{$^1$This work has been supported by Middle East Technical University Scientific Research Projects Coordination Unit under grant numbers  ADEP-704-2024-11486 and ADEP-312-2024-11455. The numerical calculations reported in this paper were  performed at TUBITAK ULAKBIM, High Performance and Grid Computing Center (TRUBA). Dr. Akbas gratefully acknowledges the support of TUBITAK 2219.} & Guillaume Picaud$^{4}$\thanks{$^4$Thanks to the ANRT and Cicat-Occitanie for funding.} \\
    Marc Chaumont$^{4}$ & Gérard Subsol$^{4}$ & Luc Téot$^{6}$ & Fahad Alsharekh$^{7}$ \\
    Shahad Alghannam$^{7}$ & Hexiang Mao$^{8}$ & Wenhua Zhang$^{9}$ & \\
  \end{tabular}
}

\address{
$^1$Middle East Technical University, Ankara 06800, Türkiye \qquad 
$^2$Helmholtz Munich, Germany \\ 
$^3$Bio-totem Pte Ltd, Suzhou, P.R. China \qquad 
$^4$LIRMM, équipe ICAR, Univ. Montpellier, CNRS, France\\ 
$^5$Univ. Nîmes, Place Gabriel Péri, France \qquad 
$^6$Cicat-Occitanie, Montpellier, France \\
$^7$Thiqah Business Services Riyadh, Saudi Arabia \qquad 
$^9$Shanghai University Shanghai 200444, China \\
$^8$Nanjing Tech University, Nanjing, Jiangsu 211816, China
}

\title{Colorectal Cancer Tumor Grade Segmentation in Digital Histopathology Images: From Giga to Mini Challenge}

\begin{document}
\maketitle
\thispagestyle{firststyle}
\begin{abstract}
Colorectal cancer (CRC) is the third most diagnosed cancer and the second leading cause of cancer-related death worldwide. Accurate histopathological grading of CRC is essential for prognosis and treatment planning but remains a subjective process prone to observer variability and limited by global shortages of trained pathologists. To promote automated and standardized solutions, we organized the ICIP Grand Challenge on Colorectal Cancer Tumor Grading and Segmentation using the publicly available METU CCTGS dataset. The dataset comprises 103 whole-slide images with expert pixel-level annotations for five tissue classes. Participants submitted segmentation masks via Codalab, evaluated using metrics such as macro F-score and mIoU. Among 39 participating teams, six outperformed the Swin Transformer baseline (62.92 F-score). This paper presents an overview of the challenge, dataset, and the top-performing methods.
\end{abstract}

\section{Introduction}

\label{sec:intro}

Colorectal cancer (CRC) represents the third most frequently diagnosed malignancy globally, accounting for over 1.8 million new cases annually \cite{sung2021global}, and stands as the second leading cause of cancer-related mortality worldwide \cite{siegel2020colorectal}. Epidemiological projections anticipate a significant increase in incidence, with an estimated 3.2 million new CRC cases expected by 2043 \cite{xi2021global}. The disease is characterized by considerable pathophysiological heterogeneity, encompassing multiple histological subtypes, each with distinct prognostic and therapeutic implications \cite{guinney2015consensus}. 
Histopathological evaluation of tissue specimens obtained via colonoscopy or surgical resection remains the gold standard for diagnosis. 
Distinguishing benign and malignant neoplasms and accurately determining tumor grade are essential components of routine pathological assessment. 
Tumor grade holds well-established prognostic significance, with poor differentiation strongly associated with adverse clinical outcomes, and  plays a critical role in guiding therapeutic decision-making \cite{cho2009histological,chandler2008interobserver,amin2017eighth}.

Despite its diagnostic relevance, histopathological grading is inherently subjective and prone to both inter- and intra-observer variability, often influenced by the level of expertise and experience of individual pathologists \cite{klein2021artificial,thomas1983observer}. This variability is further exacerbated by significant global disparities in the availability of trained pathology professionals, with some regions reporting fewer than three pathologists per million inhabitants \cite{bychkov2023constant}. These challenges highlight the urgent need for scalable, automated solutions to support and standardize diagnostic workflows. 

To address this need, we organized the ICIP Grand Challenge on Colorectal Cancer Tumor Grading and Segmentation\footnote{\url{https://sites.google.com/view/cctgs-challenge}}. The challenge utilized the METU CCTGS dataset \cite{darslan2025}, a recent, publicly available dataset. It consists of 103 whole-slide histopathology images (WSIs) from 103 patients, acquired at varying magnifications. Expert pathologists provided pixelwise annotations for five classes: tumor grades 1–3, normal mucosa, and others. The dataset includes both original high-resolution SVS files and their downsized versions. Further details on the dataset can be found in \cite{darslan2025}. 

We partitioned the dataset into training (70\%), validation (15\%), and testing (15\%) subsets using stratified random sampling to maintain class distribution balance at the pixel level. Both SVS files and downsized images for all three splits were provided to participants. Only the labels for the training and validation sets were shared. Participants submitted predicted segmentation masks to an evaluation server on Codalab\footnote{https://codalab.lisn.upsaclay.fr/competitions/22064}, which scored the submissions using the hidden test labels and returned performance metrics including F-score, precision, recall, and mean Intersection-over-Union (mIoU).

A total of 39 teams participated in the challenge. To ensure reproducibility, participants were required to submit Docker files along with their code. We verified that the reported results could be reproduced using these Docker environments and reviewed the submitted code to ensure that no inappropriate machine learning practices were used. Submissions were ranked based on the macro F-score across the five classes. From those who submitted the docker files, only six teams outperformed the $62.92$ macro F-score baseline obtained by Swin Transformer \cite{darslan2025}. In the following sections, we summarize these top six methods and evaluate the results.

\vspace{-2mm}
\section{Methods}
\vspace{-2mm}
\label{sec:methods}

This section summarizes the top six  methods, in the order of decreasing performance. The methods range from novel combinations of different models to careful fine-tuning of already existing models. Notably, only the top two teams utilized the full-resolution whole-slide images (i.e., large SVS files), and their performance was significantly better than the rest. Additionally, the top three teams employed ensembles of several models through voting. Before this challenge, the best performance on the METU CCTGS dataset was $62.92$ macro F-score; at the end of the challenge, the best score is now $70.2$.  






\vspace{-3mm}
\subsection{VAN+UperNet\protect\footnote{\textit{Contributed by He Bingdou and Xu Shuoyu.}}}
\vspace{-1mm}
\textbf{Model Architecture.} 
The method uses the UperNet architecture \cite{xiao2018unified} with a Visual Attention Network (VAN) \cite{guo2023visual} backbone. Training was initialized using ImageNet-pretrained weights. An auxiliary Fully Convolutional Network head \cite{long2015fully} was added, and both heads contributed to the total loss via separate loss calculations.

\textbf{Training:} Three models were trained via three-fold cross-validation. During training 512x512 crops were taken from the whole slide images using a active cropping method to sample regions based on label distributions. Each WSI was divided into a dense grid of candidate regions, with each patch scored according to the class-weighted sum of its pixels, allowing the system to sample patches proportionally to their semantic richness. This selection provides better coverage of rare classes and as a result it's particularly well-suited for histopathology, where foreground areas are sparse relative to background. Each resulting crop was then normalized and augmentations such as random rotations within the range of -45 to +45, random 90 counterclockwise rotations (1 to 3 times), symmetric flipping, color perturbations in the HSV space, and random scaling within a factor range of [1.0, 1.5] were applied.

A smaller learning rate was applied to the encoder than the decoder, 0.0001 and 0.001, respectively. The loss function used combines Dice loss and cross-entropy loss, incorporating both label smoothing and maximal restriction strategies in the loss calculation. The "background" class was excluded from loss computation. The Adam optimizer with a cosine warm-up schedule was used, with a batch size of 12, for a total of 40 epochs. The best-performing model was selected based on the highest average DICE score on the validation set.

\textbf{Inference: }
For inference, each WSI was split into 1024×1024 overlapping patches (50\% overlap). Patches were segmented independently by three models; their outputs were multiplied with a Gaussian kernel to emphasize central regions. The resulting class probability maps were accumulated and normalized to produce a unified per-pixel probability tensor. Final heatmaps were obtained by averaging predictions from all three models, and the final segmentation masks were produced by assigning each pixel to the class with the highest predicted score. These masks were then resized to the test set dimensions using nearest-neighbor interpolation, ensuring that discrete class labels were preserved. Finally, postprocessing was performed on the resized mask: for each class of interest, external contours were extracted and re-drawn to fill small holes and correct fragmented shapes.

\vspace{-3mm}
\subsection{DPT+MaxViT\protect\footnote{\textit{Contributed by Ümit Mert Çağlar and Alptekin Temizel.}}}
\vspace{-1mm}

\textbf{Model Architecture:} 
A set of multiple Dense Prediction Transformer (DPT) models  \cite{Ranftl2020} and Multi-Axis Vision Transformer (MaxViT) \cite{tu2022maxvit} encoders of varying sizes are trained to ensure strong generalization to real test data. Models in this set used different input scalings of data, different patch sizes, strides and loss functions. 

\textbf{Training: }
Crops of size 2048×2048 were taken out of whole slide images and then downscaled. Adaptive augmentation policy optimization with LLM feedback \cite{duru2024adaptive} was used in training, enabling the adaption of various models and configurations to the same dataset and task. It was shown that optimizing augmentation policy adaptively improved the overall training performance and robustness of the trained models.

\textbf{Inference: }
For model output handling, nearest-neighbor interpolation was used when downscaling segmentation masks to preserve class labels, while bilinear and area interpolation were also applied depending on the target. Area interpolation was particularly useful for per-pixel probability matrices, commonly applied in topographical analysis and well-suited for histopathology WSIs due to its ability to maintain regional structures \cite{roszkowiak2017survey}.

Initially, a hard-voting ensemble was applied by majority voting over the predicted masks. This straightforward approach improved the top-performing model’s mean F1 score from 67.11 to 69.07. To further improve segmentation performance, a soft voting ensemble strategy was implemented. Instead of voting on the discrete segmentation masks, the class-wise probability distributions output by each model were aggregated. For each pixel location, the predicted top-N probabilities across a selected best-performing M models were summed, and a soft class-specific bias vector was applied. This bias vector favored tumor-related classes and reduced the influence of the background predictions. 
Final masks were obtained by assigning each pixel to the class with the highest probability in the aggregated tensor. This Top-N Soft Biased Voting method achieved another boost, raising the mean F1 score to 69.73.

To refine the output of the ensemble segmentation masks and further reduce prediction noise, a three-stage postprocessing pipeline was applied. This final refinement step helped eliminate minor artifacts and boosted the final mean F1 score to 69.84 on the test set. A Gaussian filter was applied to the ensembled prediction probability maps to spatially smooth the probabilities before thresholding. This helped integrate contextual information around each pixel, mitigating sharp and isolated errors often observed at tumor boundaries. The binarized segmentation masks were refined using morphological closing operations (dilation followed by erosion) to fill small holes and close narrow gaps, improving the shape integrity of predicted tumor regions and aligning better with histopathological patterns. Small spurious components were removed by discarding connected regions with area below a dataset-specific threshold $\epsilon$, as these were typically noise or false positives introduced by edge uncertainty.

\vspace{-3mm}
\subsection{HardNet+Lawin\protect\footnote{\textit{Contributed by Guillaume Picaud, Marc Chaumont, Gérard Subsol and Luc Téot.}}}
\vspace{-1mm}

\begin{figure}
    \centering
    \includegraphics[width=.9\linewidth]{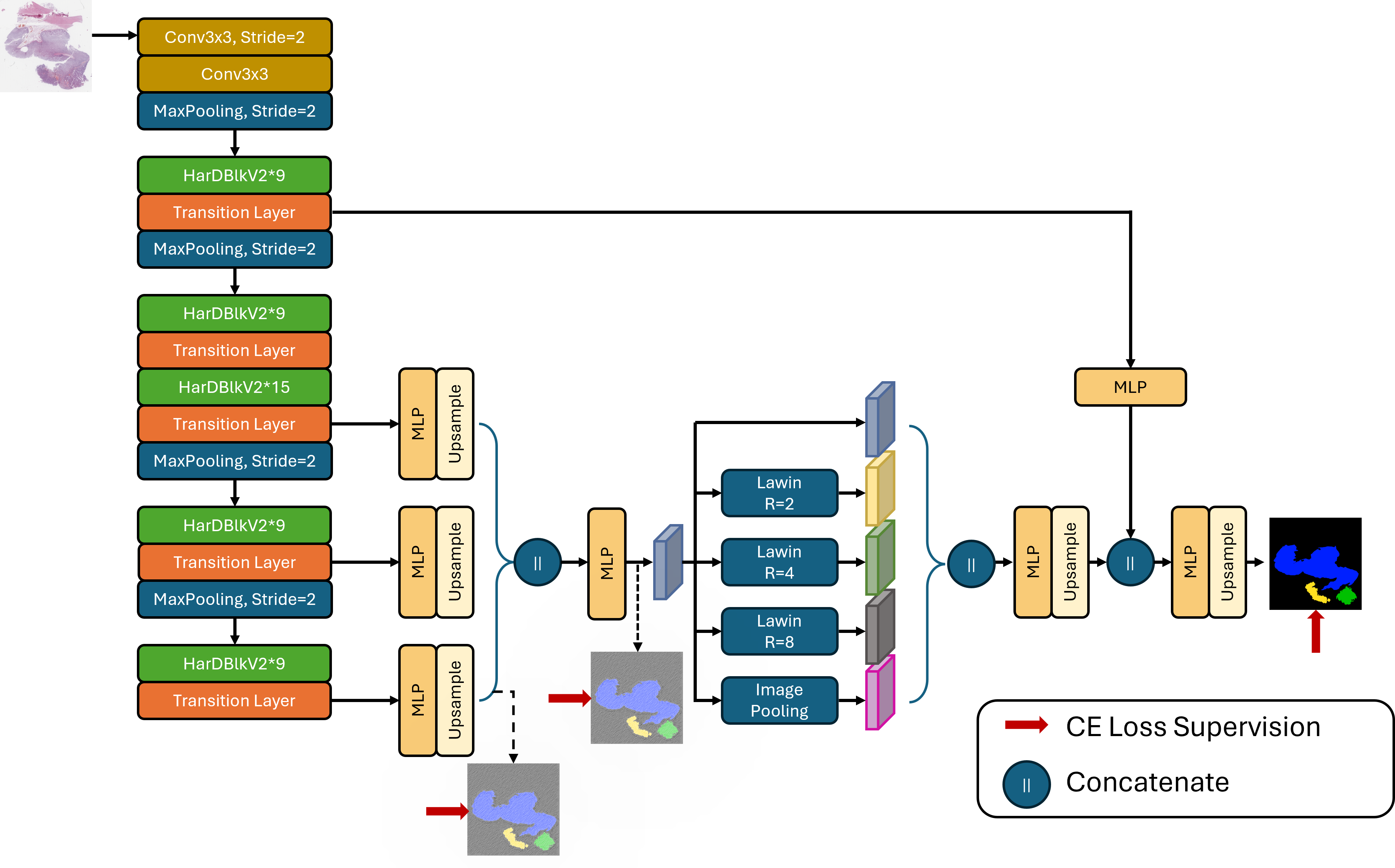}
    \caption{Model architecture for the ``HardNet+Lawin" method.}
    \label{fig:hardnet}
\end{figure}

\paragraph*{Model Architecture} 
HarDNet-DFUS \cite{HarDNet-DFUS} is a segmentation model that 
combines a lightweight convolutional encoder, HarDNet \cite{HarDNet}, with a transformer-based decoder, Lawin \cite{yan2023lawin}. Feature maps are extracted at four levels of the encoder, enabling the decoder to exploit multi-scale representations. The final three feature maps are passed through MLP layers, concatenated, and then fed into the Lawin modules \cite{yan2023lawin}. The first feature map is processed through an MLP and injected at the end of the decoder, helping preserve spatial information. The overall model architecture is illustrated in Figure \ref{fig:hardnet}. 

\textbf{Training: }
Firstly, the HarDNet encoder was retrained for tissue patch classification using the NCT-CRC-HE-100K-NONORM dataset \cite{NCT-CRC-HE-100K}. A classification head was appended to the encoder and trained by cross-entropy loss. Although the pretraining dataset has known biases~\cite{CRCBiais}, the pretrained weights sped up convergence and boosted early validation performance. Three unfreezing schedules, fully frozen encoder, warm up, and progressive unfreezing, were compared against one and another and it was found that, the warm-up strategy provided the most consistent validation improvements. 
 
 Weights learned from the pretraining step were used to initialize the encoder part of the HarDNet-DFUS model for the segmentation task. Cross-entropy loss function and the AdamW optimizer with an initial learning rate of 1e-4 (scheduled via cosine annealing) and an Exponential Moving Average of the weights were employed during training. Downsampled version of the CCTGS dataset was used. To mitigate catastrophic forgetting, a warm-up phase of two epochs at the beginning of training was included \cite{catastrophicforgetting}. Images were resized to 1536×1536 pixels and batch size of two was selected due to memory limitations. A standard data augmentation pipeline, combining geometric transformations including flipping, shifting, scaling and rotation and color transformations including color jittering, coarse dropout and Gaussian Noise was applied. The proposed training strategy was applied using 5-fold cross-validation, with the mean F1-score as the validation metric.

\textbf{Inference: }
To address performance variability across folds, an ensembling strategy by selecting the top three models, applying test-time augmentation (TTA) independently to each, and aggregating their predictions through majority voting was adapted. The final predictions were post-processed using hole filling.
\vspace{-3mm}

\vspace{-3mm}
\subsection{Segmenter-L\protect\footnote{\textit{Contributed by Fahad Alsharekh.}}}
\vspace{-1mm}
This method basically fine-tunes a pre-trained, off-the-shelf Segmenter-L with ViT backbone model from MMSegmentation \cite{mmseg2020}. The model was pre-trained on the ADE20K dataset. 

\vspace{-6mm}
\subsection{SegFormer-B1\protect\footnote{\textit{Contributed by Shahad Alghannam.}}}
\vspace{-1mm}
This method basically fine-tunes a pre-trained, off-the-shelf SegFormer-B1 model from the MMSegmentation repository \cite{mmseg2020}. The model was pre-trained on the Cityscapes dataset.

\vspace{-3mm}
\subsection{PathVTA\protect\footnote{\textit{Contributed by Hexiang Mao and Wenhua Zhang.}}}
\vspace{-1mm}



\def\x{{\mathbf x}}
\def\L{{\cal L}}

%

\textbf{Model Architecture:} 
It consists of three key components: (1) the UNI\cite{chen2024uni} foundation model serving as the feature extractor, (2) a ViT Adapter\cite{chen2022vitadapter} module that yields a multi-scale feature pyramid, and (3) a UNet-style decoder that restores spatial resolutions and predicts the segmentation mask. The UNI backbone is frozen, and only the adapter and decoder are trained. 

\textbf{Training: }  
PathVTA was trained on the down-scaled dataset. To improve data diversity, augmentation techniques such as random cropping, rotations and flips as well as color jittering, gaussian blur and photometric distortion were employed. Training was performed for 5,000 iterations using cross-entropy loss. Adam optimizer with an initial learning rate of 0.0002 was used to train the network.

\textbf{Inference: } 
To handle high-resolution inputs, a sliding window inference strategy with a window size of (224, 224) and a stride of (112, 112) employed. For each extracted patch, the model outputs logits, which are converted to class probabilities via softmax. For overlapping regions, predicted probabilities are accumulated. Then, a second softmax is applied to normalize the aggregated probability map. Finally, the segmentation prediction is obtained by performing the argmax over the normalized map.

\vspace{-5mm}
\section{Results}
\vspace{-2mm}
\begin{table*}[h!]
  \centering
  \caption{Performances of top six methods and the baseline.}
  \label{tab:methods}
  \setlength{\tabcolsep}{3pt} 
  \begin{tabular}{cccccc}
    \toprule
    \textbf{Method}
    
    & \textbf{Notes}  
      & \textbf{mFscore} 
      & \textbf{mIoU} 
      & \textbf{mPrecision} 
      & \textbf{mRecall} \\
    \midrule
    \textbf{VAN+UperNet} & WSI \& ensemble & 70.2  & 56.5 & 67.2 & 74.4 \\
    \textbf{DPT+MaxViT} & WSI \& ensemble & 69.8 & 55.8 & 68.6 & 71.7 \\
    \textbf{HardNet+Lawin} & ensemble & 66.7  & 52.8 & 63.0 & 72.4 \\
    \textbf{Segmenter-L} & & 65.2 & 51.3 & 63.7 & 67.5 \\
    \textbf{Segformer-B1} & & 65.1 & 50.7 & 61.2 & 70.6 \\   
    \textbf{PathVTA} & & 64.2 & 50.2 & 63.3 & 65.7 \\
    \textbf{SwinTransformer} \cite{darslan2025} & baseline & 62.9 & - & 60.9	& 69.6 \\
     \bottomrule
  \end{tabular}
\end{table*}


Performances of the six contributed methods can be seen in Table  \ref{tab:methods}. 
Although the dataset included large whole-slide images (WSIs), only two teams used them—likely because handling and processing such high-resolution files is technically challenging.
However, we see a clear benefit of using WSIs on test metrics. The top two teams both utilized WSI and they are both at least 3 points above the third team. 
Utilizing features extracted from the original high-resolution whole slide images (WSIs), as opposed to their downsampled counterparts, allows models to capture richer morphological details, thereby enabling more accurate and fine-grained analysis. Another trend we see is that the methods using voting mechanism tended to perform better than others. This may be because ensembling combines the strengths of multiple models, leading to more robust and accurate predictions than single-model approaches.

\begin{figure*}
\centering
\includegraphics[width=.9\linewidth]{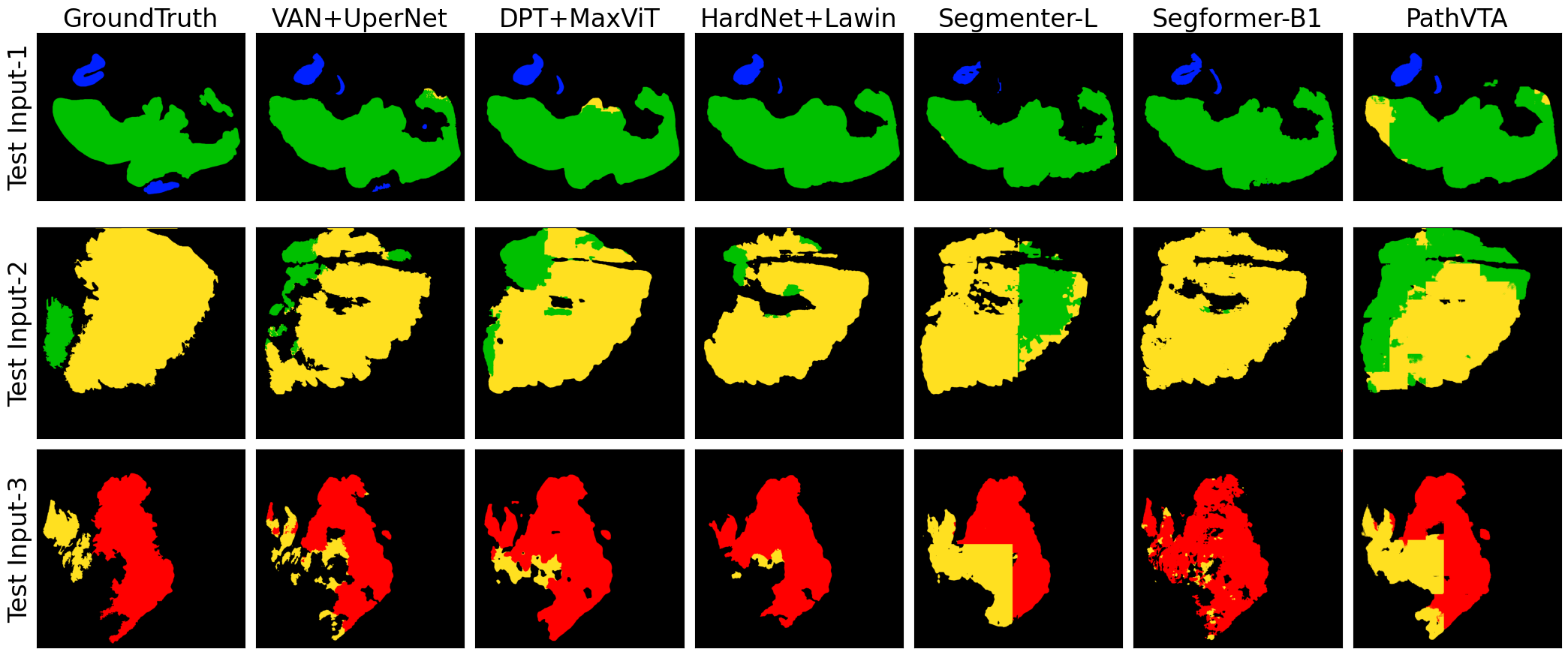}
\vspace{-3mm}
\caption{Qualitative experimental results for Top-6 best performing methods for 3 input test images. (Color codes for classes: Blue:Normal mucosa, Green:Grade-1, Yellow:Grade-2 , Red: Grade-3, Black:Others)}
\label{fig:res}
\end{figure*}

Representative qualitative results for tumor grade segmentation generated by  the top six  methods are shown in Figure~\ref{fig:res}, highlighting their comparative performance across different histopathological cases. \textbf{VAN+UperNet} demonstrates high precision in identifying semantic class boundaries, even for non-convex and non-compact objects. It successfully detects normal mucosa with minimal omission. Although it shows some confusion between Grade 1 and Grade 2 tumors, it performs well in distinguishing Grade 3 tumors, rarely misclassifying them as Grade 2 and generally separating them clearly from other tumor types. In terms of minimizing confusion between Grade 1 and Grade 2, it ranks among the top-performing methods after SegFormer-B1.  Compared to VAN+UperNet's results, \textbf{DPT+MaxViT} tends to label larger areas as Grade 3 in Grade 3-positive images. It effectively segments normal mucosa, even when boundaries are non-convex. However, it struggles with distinguishing between Grade 1 and Grade 2 tumors. Like VAN+UperNet, it often labels ambiguous “others” regions as Grade 3 in Grade 3-positive images. \textbf{HardNet+Lawin} tends to over-connect boundaries when segmenting non-convex normal mucosa regions. It shares similar issues with other methods in confusing Grade 2 and Grade 3 tumors, but demonstrates a lower tendency to detect Grade 3 regions in images that contain them, often missing these tumors.

In summary, VAN+UperNet appears to offer the most balanced performance in preserving class boundaries and minimizing confusion, particularly for complex tumor morphologies. SegFormer-B1 excels in grade separation but is limited by fragmented predictions. Segmenter-L and DPT+MaxViT exhibit strong detection for certain classes but suffer from over-segmentation and grade confusion. HardNet+Lawin underperforms in high-grade detection, while  PathVTA show extensive misclassifications and it is particularly prone to inter-grade confusion. Overall, while each method has strengths, challenges remain particularly in resolving boundary-level ambiguities and differentiating intermediate-grade tumors.
These observations reinforce the need for robust spatial modeling and grade-aware architectures in semantic segmentation for tumor analysis.

The consistency and style of training annotations could be studied. By visualizing training annotations, we distinguished two annotation methods: one produces masks with smoothed boundaries, while the other yields masks with highly detailed edges. This discrepancy introduces learning challenges, as alternating between annotation domains may lead to undesired biases. It also complicates model evaluation, since there is no clear guidance on which annotation domain should be prioritized.



\label{sec:res}

\vspace{-5mm}
\section{Conclusion}
\vspace{-2mm}
\label{sec:conc}

The ICIP Grand Challenge on Colorectal Cancer Tumor Grading and Segmentation aimed to address the pressing need for accurate models for tumor grading and segmentation in histopathology by benchmarking algorithms on a realistic dataset. The challenge attracted broad participation, with 39 teams developing segmentation models for complex multi-class tissue grading and segmentation under limited supervision. The top-performing methods demonstrated the potential of advanced deep learning models, particularly transformer-based architectures, for handling high-resolution whole-slide images with dense pixelwise annotations. Model ensembling also stood out as an important technique for improved accuracy.

While the leading approaches showed promise, the overall performance margins indicate that histopathology tumor grading and segmentation remains a difficult task, especially in discriminating between tumor grades. These findings highlight opportunities for future research, including more effective use of multi-scale information and better data efficiency in training.  The METU CCTGS dataset and the evaluation infrastructure remain publicly available, encouraging continued progress in this critical area of medical image analysis.

\vspace{-3mm}
\bibliographystyle{IEEEbib}
\bibliography{references}

\end{document}